\documentclass[letterpaper, 10 pt, conference]{ieeeconf}  % Comment this line out if you need a4paper

\IEEEoverridecommandlockouts                              % This command is only needed if 
\usepackage[per-mode=symbol]{siunitx}
\usepackage{graphics} % for pdf, bitmapped graphics files
\usepackage{epsfig} % for postscript graphics files
\usepackage{graphicx}
\usepackage{tabularx}
\usepackage{multirow}
\usepackage{multicol}
\usepackage{enumerate}
\usepackage{url}
\usepackage{subcaption}
\usepackage{dblfloatfix}
\usepackage{amsmath}
\usepackage{algorithm}
\usepackage{algpseudocode}
\usepackage{hyperref}
\title{\LARGE \bf
Watch out! There may be a Human. \\Addressing Invisible Humans  in Social Navigation
}

\author{Phani Teja Singamaneni$^{1}$, Anthony Favier$^{1,2}$, Rachid Alami$^{1,2}$ % <-this % stops a space
\thanks{$^{1}$Authors are with LAAS-CNRS, Universite de Toulouse, CNRS, Toulouse, France, {\{ptsingaman, anthony.favier, rachid.alami\}@laas.fr} }%
\thanks{$^{2}$This  work  has  been  partially  funded  by  the  Agence  Nationale  de  la Recherche through the ANITI ANR-19-PI3A-0004 grant}
}

\begin{document}

\maketitle
\thispagestyle{empty}
\pagestyle{empty}

%%%%%%%%%%%%%%%%%%%%%%%%%%%%%%%%%%%%%%%%%%%%%%%%%%%%%%%%%%%%%%%%%%%%%%%%%%%%%%%%
\begin{abstract}
% The current approaches in human-aware or social robot navigation address the humans that are visible to the robot. However, it is also important to address the possible emergences of the humans to avoid shocks or surprises to humans and erratic behavior of the robot planner. In this paper, we propose a novel approach to address such human emergences called `invisible humans'. Apart from the detection of invisible humans, we also show how this detection can be exploited to identify and address the doorways or narrow passages. We finally integrate the proposed framework into our human-aware planner, CoHAN, and show the results in several interesting scenarios.
Current approaches in human-aware or social robot navigation address the humans that are visible to the robot. However, it is also important to address the possible emergences of humans to avoid shocks or surprises to humans and erratic behavior of the robot planner. In this paper, we propose a novel approach to detect and address these human emergences called `invisible humans'. We determine the places from which a human, currently not visible to the robot, can appear suddenly and then adapt the path and speed of the robot with the anticipation of potential collisions. This is done while still considering and adapting humans present in the robot's field of view. We also show how this detection can be exploited to identify and address the doorways or narrow passages. Finally, the effectiveness of the proposed methodology is shown through several simulated and real-world experiments.
\end{abstract}

%%%%%%%%%%%%%%%%%%%%%%%%%%%%%%%%%%%%%%%%%%%%%%%%%%%%%%%%%%%%%%%%%%%%%%%%%%%%%%%%
\section{Introduction}
Human-aware or social robot navigation is rapidly advancing, and new frameworks \cite{kollmitz2015time, triebel2016spencer, repiso2017line, guldenring2020learning} are required to address the intricate social navigation scenarios. However, most of these frameworks \cite{moller2021survey, kruse_human-aware_2013} address only the visible humans and do not take into account the possible emergence of humans that are not visible currently. We believe that such invisible humans should be considered while developing a human-aware navigation framework to avoid any erratic behaviors of the robot planner when a human suddenly appears. Therefore in this work, we try to address these invisible humans in social navigation settings.

There are not many works that address this problem in the field of human-aware navigation. However, there are some existing works in classical robot navigation that address similar issues. Particularly, this work is inspired by the pioneering work of M. Krishna concerning the ability of a mobile robot, based on the model of its perception functions, to assess from where in the close environment of the robot a human can emerge and prepare to react to ensure no-collision by adapting its path and velocity \cite{madhavakrishna-ra-2006, alami-springer-2007, madhavakrishna-iros-2003}. Some recent works like \cite{chung2009safe, Bouraine-2012} address the issues of robot navigation in occluded or unknown regions with a limited field of view. The work presented in \cite{miura2006adaptive} talks about the adaptive speed control of the robot in unknown environments and also talks about the occluded regions. The authors of \cite{lambert2008collision} propose a methodology to mitigate or avoid collisions while navigating. In our case, we are trying to mitigate possible future collisions with a human. The motion planning problem in the presence of dynamic obstacles can be solved in several ways. Some works use grid-based planning \cite{kollmitz2015time, phillips2011sipp} while some others \cite{schoels2020nmpc, rosmann2021online} use optimizations like model predictive control. We use sparse graph optimization proposed in \cite{rosmann2013efficient} and add human-aware constraints as proposed in our previous works \cite{khambhaita2017viewing, singamanenihateb2020, singamaneni2021human}.

% As it is evident that the unknown or occluded region could cause issues with classical navigation, the same applies to social navigation. Hence, we propose the concept of invisible humans to human-aware navigation planning through this paper. As per the knowledge of the authors, there is no other work that addresses this problem in human-aware navigation. The major contributions of this paper are as follows. Firstly, we formulate the problem of invisible humans detection and propose an algorithm to locate them. Then, we integrate these invisible humans into our human-aware navigation framework, CoHAN \cite{singamaneni2021human}, by introducing a new human-aware constraint into our optimization scheme. We further show how the detected invisible humans can be exploited to identify some interesting scenarios and address these by adding new modalities to CoHAN. The implementation and code can be found at {\small {\url{https://github.com/sphanit/cohan_planner_multi/tree/model}}}. 
As it is evident that the unknown or occluded region could cause issues with classical navigation, the same applies to social navigation. Hence, we propose the concept of `invisible humans' to human-aware navigation planning through this paper. As per the knowledge of the authors, there is no other work that addresses this problem in human-aware navigation. The major contributions of this paper are as follows. Firstly, we formulate the problem of invisible humans detection and propose an algorithm to determine the locations from which humans who are not currently visible can emerge suddenly. These invisible humans are then integrated into our human-aware navigation framework, CoHAN \cite{singamaneni2021human}, by introducing a new human-aware constraint into our optimization scheme. This constraint modifies the path and speed of the robot taking into account the anticipation of potential human appearances to avoid collisions and surprises. We further show how the detected invisible humans can be exploited to identify some interesting scenarios (like doors/passages) and address these by adding new modalities to CoHAN. The implementation and code can be found at {\small {\url{https://github.com/sphanit/cohan_planner_multi/tree/model}}}. 

The rest of the paper is organized as follows. Section II presents the formulation and an algorithm to detect invisible humans. Section III shows how the invisible humans are integrated into CoHAN and talks about the issues that arise. It also presents a simple formulation to identify narrow passages. In Section IV, various experiments to evaluate the proposed approach are presented, followed by the real-world experiments in Section V and a discussion on the limitations in Section VI. Finally, Section VII concludes the work.

\section{Invisible Humans Detection}
The invisible humans are detected using an emulated laser scan on a 2D map in ROS. A custom laser scan is attached to the robot's base and it is continuously updated as the robot moves on a given map. The entire system is implemented in ROS \cite{quigley2009ros} and requires the map that is published by the ROS Navigation stack. In order to avoid too many detections, we limit the invisible humans detection to a radius of \SI{5}{\meter} in front of the robot. The detection of invisible humans is a two-step process involving corner detection and locating invisible humans. Each step is explained in detail below.
\begin{figure}[h!]
    \centering
    \begin{subfigure}[t]{0.45\columnwidth}
    \centering
  \includegraphics[width=\columnwidth]{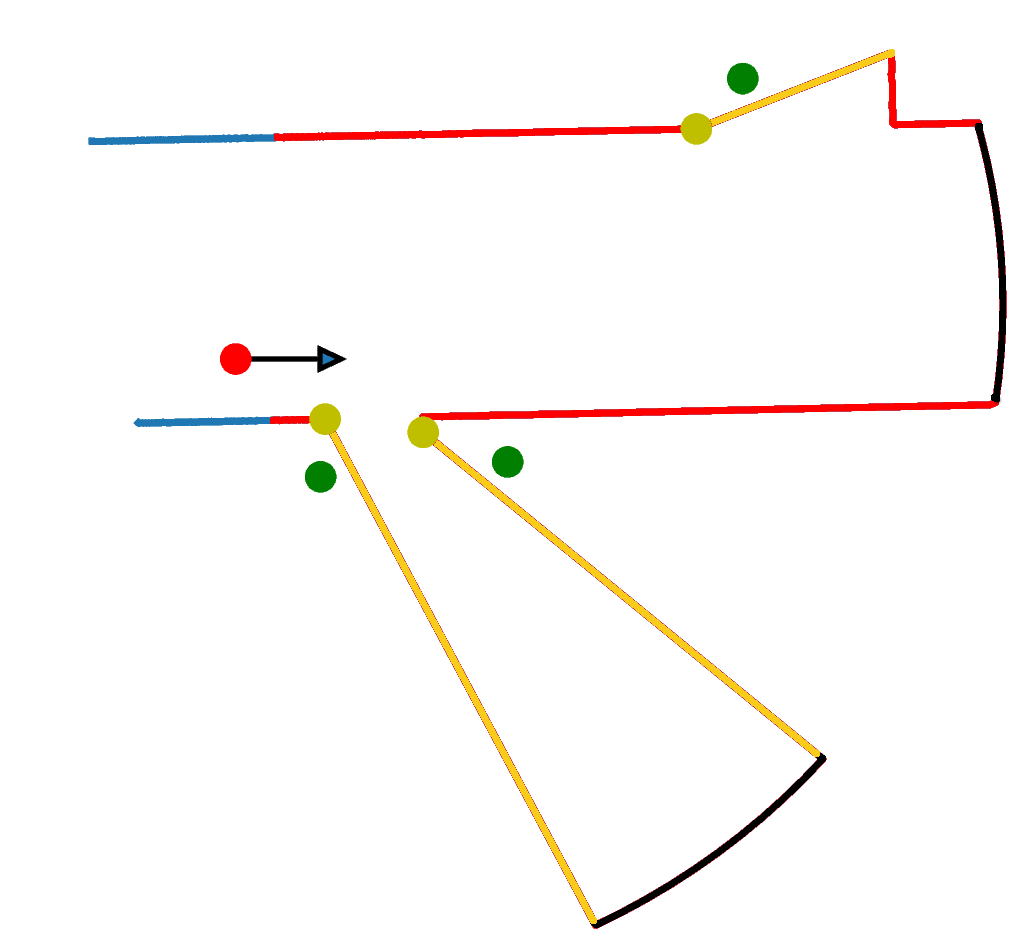}
  \caption{Laser Contour}
\end{subfigure}
\begin{subfigure}[t]{0.5\columnwidth}
\centering
  \includegraphics[width=0.8\textwidth]{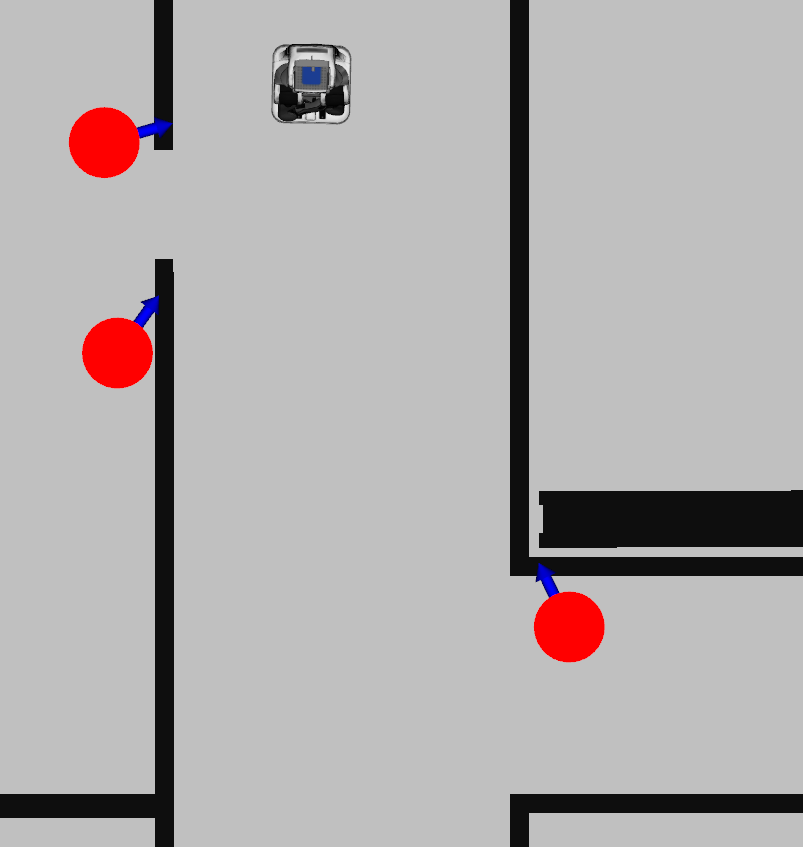} 
  \caption{Detected Invisible Humans}
\end{subfigure}
    \caption{(a) Laser contour built by the custom laser sensor. The red and blue lines are the actual walls or obstacles in the front and back of the robot respectively. The black lines are the laser range boundaries and the yellow lines are interpolated lines between two gaps of laser data. The robot is shown as a red dot with an arrow. The detected corners are shown in yellow, while the detected invisible humans are shown in green. (b) The detected invisible humans on the map for the contour shown. The red circles shows the location and the blue arrow shows the assumed direction, which is always oriented towards the robot.}
    \label{fig:detection}
\end{figure}

\subsection{Laser Contour and Corner Detection}
A custom laser scan sensor that is attached to the robot's base, scans the given 2D map to get the visible contour of the map. The laser data consists of a list of values showing the scan ranges in the field of vision of the sensor. This custom sensor data is used in the place of the real laser scan data to ensure uniformity across different robots and sensors. An example laser contour built using this is shown in Fig. \ref{fig:detection}. Different parts of these contour lines are shown in different colors for ease of explanation. The red and the blue lines together constitute the regions on the real map where the laser has hit a wall or an obstacle. The black lines represent the laser data that did not hit anything and reached the end of their range (in our case, range of the laser is \SI{7}{\meter}). Lastly, the yellow lines are interpolated rays joining large separations between consecutive laser values and play a major role in our algorithm. The red circle and the arrow represent the robot's position and its direction, while the green circles represent the estimated invisible humans. The corner detection is relatively easy once the laser contour is available. Firstly, all the pairs of consecutive laser ranges that are separated by more than \SI{0.5}{\meter} are determined and stored in a set \{$V$\}. The threshold of \SI{0.5}{\meter} is chosen to filter out small gaps from where a human will not emerge. After this, the values that are closest to the robot in each of the above pairs are identified to be corners and stored in a set \{$c$\}. These are shown as yellow circles in Fig. \ref{fig:detection}.

\subsection{Estimation of Invisible Humans}
\begin{figure}[h!]
% \centering
\begin{subfigure}{.6\columnwidth}
  % include first image
  \includegraphics[width=0.8\textwidth]{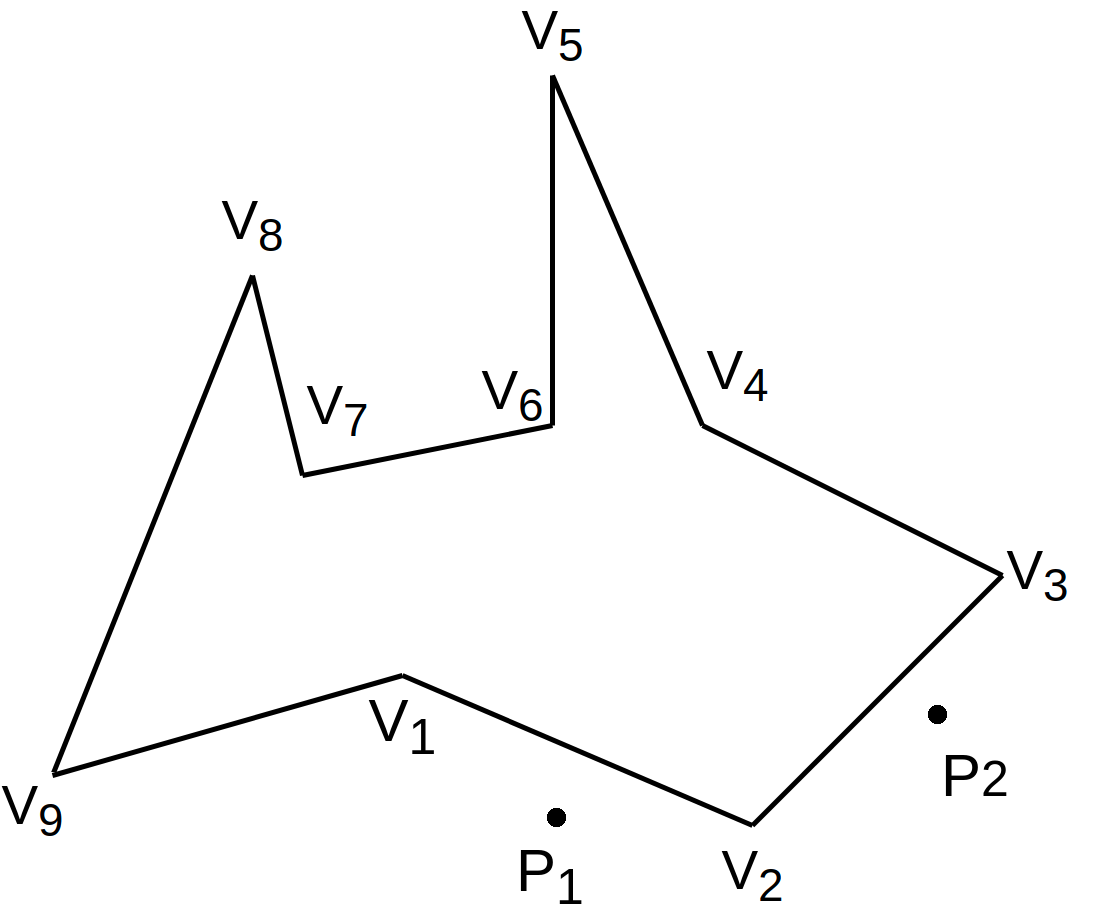}
\end{subfigure}
\hspace{-0.17cm}
\begin{subfigure}{.4\columnwidth}
%   \centering
  % include second image
  \includegraphics[width=0.8\textwidth]{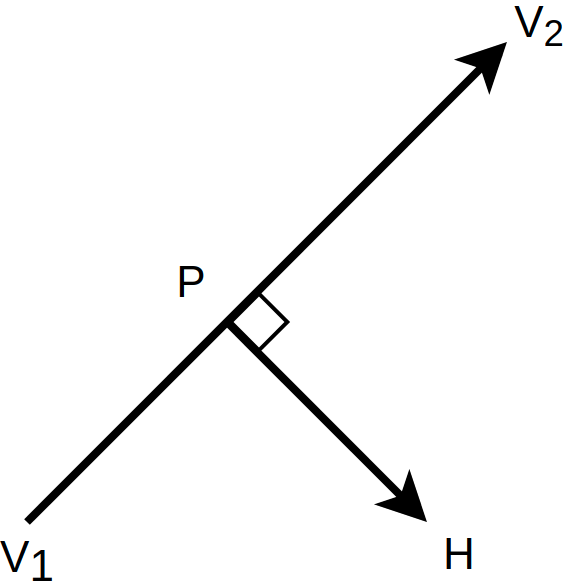} 
\end{subfigure}
\caption{Non-convex polygon and vector formulation to determine invisible humans. We try to find a point $H$ that lies between two vertices $V_1$ and $V_2$ and lies to the right of $\protect\overrightarrow{V_1V_2}$. Note that the perpendicular distance of this point should be greater than the assumed human radius.}
\label{fig:polyg}
\end{figure}

\begin{algorithm}
\caption{Locate Invisible Humans}\label{algo}
\begin{algorithmic}[1]
% \Function{FindMeshIndex}{$\vars{position}, \vars{nGrid}$}
\State Determine the vertex pairs set \{$V$\} using laser contour
\State Determine the corners set \{$c$\} from \{$V$\}
\For {each $c$} 
\State $V_1$ = $c$ = $(x_1,y_1)$
\State $V_2$ = $(x_2, y_2)$ \Comment{Corresponding pair from \{$V$\}}
\State $u_x = \frac{(x_2-x_1)}{\lVert\overrightarrow{V_1V_2}\rVert}$, $u_y = \frac{(y_2-y_1)}{\lVert\overrightarrow{V_1V_2}\rVert}$ 
\State Set $P$ = $(x_p,y_p)$ = $(x_1,y_1)$
\While{True}
\State Calculate $H_{inv}$ using Eq. (4)
\State Calculate $\overrightarrow{r}$ and $\beta$
\If{$\lVert\overrightarrow{r}\rVert < LaserData(\beta)$}
\State $x_p = x_p + \alpha u_x$
\State $y_p = y_p + \alpha u_y$
\State continue
\EndIf
\State Check for overlap on the Map
\If{no overlap}
\State $advance = False$
\For{$i = 1$ to $k$}
\State $d = \frac{i}{k}* (h_{rad}+\epsilon)$
\State Calculate $P_r$ and $P_l$ using Eqs. (4), (5)
\State Check $P_r$, $P_l$ for the overlap on Map 
\If{no overlap}
\State continue
\ElsIf{overlap}
\State $advance = True$
\State break
\EndIf
\EndFor
\EndIf
\If{$advance == True$}
\State $x_p = x_p + \alpha u_x$
\State $y_p = y_p + \alpha u_y$
\ElsIf{$advance == False$}
\State break
\EndIf
\EndWhile
\State Add $H_{inv}$ to the set of invisible humans, \{$H_{inv}$\}
\EndFor
\State \textbf{return} \{$H_{inv}$\}
\end{algorithmic}
\end{algorithm}
The estimation of possible locations for the invisible humans is not very straightforward. The laser contour forms a complex non-convex polygon, and we are searching for circles whose centers are outside this polygon and do not intersect the contour. We solve this problem using a combination of ray tracing and vector algebra. Consider a non-convex polygon as shown in Fig. \ref{fig:polyg}. The vertices are numbered in the anti-clockwise direction. Consider a point $P_1$ that lies between the vertices $V_1$ and $V_2$. If $P_1$ is outside the polygon it should lie to the right of the vector $\overrightarrow{V_1V_2}$. Similarly, a point $P_2$ lying outside the polygon between $V_2$ and $V_3$ lies to the right of $\overrightarrow{V_2V_3}$ and so on. It holds true irrespective of the number of sides of the polygon. We exploit this property to determine the positions of the invisible humans. Note that, a point lying outside the polygon will always be on the right side of the vectors, but not every point on the right of the vectors lies outside the polygon. This is because this methodology uses only a single side and does not consider the other sides. To handle this, we use the fact that the polygon in our case is determined by the laser contour and any point outside this polygon is not visible to the laser.

In Fig. \ref{fig:detection}, the yellow lines correspond to the edges of interest in the polygon. The numbering of the vertices is determined based on the indices of the laser scan data which (the rays) move from right to left in the anti-clockwise direction. In order to determine a possible invisible human, we need to determine a point $H$ that is to the right of $\overrightarrow{V_1V_2}$ and whose perpendicular distance is greater than an average human radius, $h_{rad}$ as shown in Fig. \ref{fig:polyg}. Consider a point $P$ that lies on line segment $\overline{V_1V_2}$ such that $\overrightarrow{PH}$ is perpendicular to $\overrightarrow{V_1V_2}$. If we know the point $P$, then $H$ can be determined using the following equations:

\begin{equation}
       sign(\overrightarrow{V_1V_2}\times\overrightarrow{PH}) = -1
\end{equation}
\begin{equation}
    \overrightarrow{V_1V_2}\cdot\overrightarrow{PH} = 0
\end{equation}
\begin{equation}
           \lVert\overrightarrow{PH}\rVert = h_{rad}+\epsilon
\end{equation}
where ($\times$) is the cross product, ($\cdot$) is the dot product and ($\lVert \rVert$) is the euclidean norm respectively. $\epsilon$ is the off-set on the human radius to avoid unrealistic detections. In this work, we chose $\epsilon$ to be $\frac{h_{rad}}{2}$. Let $V_1$ = $(x_1, y_1)$, $V_2$ = $(x_2, y_2)$ and $P$ = $(x_p, y_p)$, then by solving the above equations, $H$ = $(x_{h}, y_{h})$ is given by:
\begin{equation}
  \begin{aligned}
        x_{h} = x_p + \frac{d(y_2-y_1)}{\sqrt{(x_2-x1)^2+(y_2-y1)^2}} \\
        y_{h} = y_p - \frac{d(x_2-x_1)}{\sqrt{(x_2-x1)^2+(y_2-y1)^2}}  
  \end{aligned}
\end{equation}
where $d = (h_{rad}+\epsilon)$. Similarly, the point on the left can be obtained by reversing the sign in Eq. (1), which yields: 
\begin{equation}
  \begin{aligned}
        x_{h} = x_p - \frac{d(y_2-y_1)}{\sqrt{(x_2-x1)^2+(y_2-y1)^2}} \\
        y_{h} = y_p + \frac{d(x_2-x_1)}{\sqrt{(x_2-x1)^2+(y_2-y1)^2}}  
  \end{aligned}
\end{equation}

From the Eq. (4), we can see that $(x_p,y_p)$ is required to determine $(x_h,y_h)$ and it is still unknown as we cannot solve for four variables using only three equations (Eq. (1)-(3)). As it is already known that $P$ lies on the line segment joining $\overline{V_1V_2}$, it can be determined by performing a search on this line segment, starting at one end and moving towards the other in small increments. The set of detected corners, \{$c$\} are taken as the starting points of this search. In each iteration, a possible invisible human position is estimated using Eq. (4) and then projected onto the map to see if there is any overlap with an obstacle or wall. As mentioned before, we need another check to ensure that the point is outside the laser contour. Suppose, the vector joining the robot and the point $H$ is $\overrightarrow{r}$ and it subtends an angle $\beta$ with the positive $x-axis$ of the base frame of the robot. As the custom laser is also attached to the base frame of the robot, there should be laser scan data corresponding to this angle $\beta$. Hence, when the $H$ is outside, the following condition is satisfied:
\begin{equation}
    \lVert\overrightarrow{r}\rVert > \rho(\beta)
\end{equation}
where $\beta = atan2(x_h - x_{rb}, y_h-y_{rb})$, $(x_{rb}, y_{rb})$ is the robot base frame's position and $\rho(\beta)$ is the laser scan reading at angle $\beta$. To refine this search further, two points, one on left, $P_l$, and the other on the right, $P_r$, of the $H$ are considered with incremental distances until $h_{rad}+\epsilon$ and checked for overlap using the map. 

The entire procedure is shown below in Algorithm \ref{algo} where $u_x$ and $u_y$ (line 6) are the unit vectors along the direction of $\overrightarrow{V_1V_2}$ and $\alpha$ is a scalar determining (lines 26, 27) the step size or increment.
% \begin{figure}
%     \centering
%     \includegraphics[width=0.7\columnwidth]{figures/detection_rviz.png}
%     \caption{The detected invisible humans on the map for the situation shown in Fig. \ref{fig:detection}. The red circles shows the location and the blue arrow shows the assumed direction, which is always oriented towards the robot.}
%     \label{fig:rviz_detect}
% \end{figure}
The invisible humans detected using the above-mentioned algorithm are shown in Fig. \ref{fig:detection} (b). The red circles are the detected location while the blue arrows show the direction. We assume that the humans are always coming towards the robot and hence, the direction is always oriented towards the robot. In the next section, we explain how this is integrated into the human-aware planning framework for social robot navigation.  
%% Useful for Subfigures
% \begin{figure}[h!]
% \begin{subfigure}{.5\columnwidth}
% %   \centering
%   % include first image
%   \includegraphics[width=\textwidth]{figures/detection.png}
% \end{subfigure}
% \hspace{-0.17cm}
% \begin{subfigure}{.5\columnwidth}
% %   \centering
%   % include second image
%   \includegraphics[width=\textwidth]{figures/detection_rviz.png} 
% \end{subfigure}
% \caption{Detection and laser contour}
% \end{figure} 

\section{Integration with a Human-Aware Planner}
In the previous version of CoHAN, we address different types of visible humans by introducing new modalities and human-aware constraints \cite{singamaneni2021human}. In this work, we extend it further to address the invisible humans. The invisible humans are detected as explained above and then they are published on a ROS topic. CoHAN subscribes to this topic and adds a new constraint to its optimization that is specifically designed for invisible humans. Further, using these invisible human detections, we propose a methodology to identify doors and narrow passages. For more details on the modality switching and constraint implementation readers are advised to refer to our previous works \cite{ khambhaita2017viewing, singamanenihateb2020, singamaneni2021human}.
\subsection{Invisible Humans Constraint}
The invisible humans constraint takes into account the human reaction time, walking speed, and deceleration and aims to make the robot cautious about the sudden human emergence. The cost added by this constraint for the n\textsuperscript{th} pose of the robot's trajectory is given as:
\begin{equation}
\begin{split}
cost_{inv\_human} &= max\left(\frac{V-a\Delta t_n}{d}, 0\right)\quad if\quad \Delta t_n> 0.5s\\
                &= \frac{V}{d}\quad otherwise
\end{split}
\label{cost_}
\end{equation}
% where $d$ is the distance between the invisible human and the robot, $V$ is the average human walking speed, \SI{1.3}{\meter\per\second} \cite{phdthesis} and $a$ is the deceleration of the human. The value of this deceleration can vary and can be up to a maximum of \SI{2.94}{\meter\per\second^2} (\SI{0.3}{\g}) \cite{lakoba2005modifications}. In this work we take a reaction time of \SI{0.5}{\second} as discussed in \cite{helbing2000simulating,lakoba2005modifications}. Hence the constraint adds maximum possible cost until \SI{0.5}{\second}. Then we assume that the human will continuously decelerate to avoid collision with the robot over time and eventually stops which is reflected in the top part of Eq. \eqref{cost_}. The time and human detections are reset after every control cycle.
where $d$ is the distance between the invisible human and the robot, $V$ is the average human walking speed, \SI{1.3}{\meter\per\second} \cite{phdthesis}, $a$ is the deceleration of the human and $\Delta t_n$ is the time difference between the n\textsuperscript{th} pose and the starting pose of the planned trajectory of the robot. The value of the deceleration, $a$, can vary and can be up to a maximum of \SI{2.94}{\meter\per\second^2} (\SI{0.3}{\g}) \cite{lakoba2005modifications}. In this work we take a reaction time of \SI{0.5}{\second} as discussed in \cite{lakoba2005modifications, helbing2000simulating}. Hence the constraint adds the maximum possible cost until \SI{0.5}{\second}. Then we assume that the human will continuously decelerate to avoid collision with the robot over time and eventually stops, which is reflected in the upper part of Eq. \eqref{cost_}. The time ($\Delta t$) and human detections are reset after every control cycle.
\subsubsection{Issue with the constraint}
The main objective of the constraint is to push the robot away from the opening, anticipating the emergence of invisible humans. However, when the robot needs to pass through this opening and if the passage is narrow (door or narrow corridor), the constraint pushes the robot away and makes it impossible to enter the passage. To mitigate this, we devise a simple formulation that detects such scenarios. Once a narrow passage is detected, the invisible humans constraint is switched off, and the maximum robot's velocity is reduced until it passes through. The passage detection process is explained in detail below.
\subsection{Doorway or Narrow Passage Detection}
\begin{figure}[h]
    \centering
    \includegraphics[width=0.8\columnwidth]{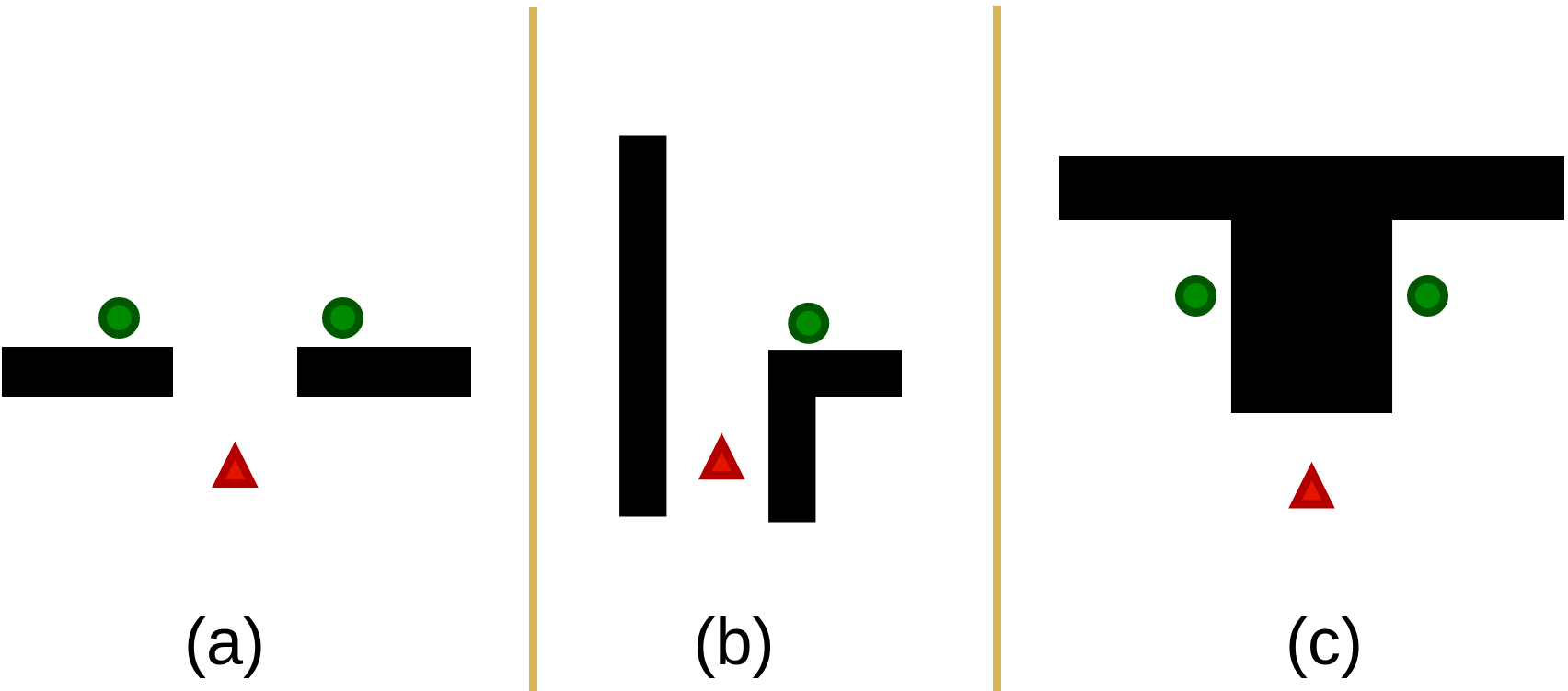}
    \caption{Different types of passages that are detected using invisible humans. (a) Doorways or openings and endings of the corridors (b) a narrow passage with opening on one side and a wall on the other side (c) a large pillar or obstacle where robot cannot see on either side. The green circles are the possible locations of the invisible humans and the red triangle shows the robot pointed towards the direction of its motion.}
    \label{fig:passages}
\end{figure}

The detection of narrow passages or doorways not only allows us to overcome the issue of the invisible humans constraint but also allows us to define a new modality of planning that needs to be handled separately. In this work, we try to address three different scenarios, as shown in Fig. \ref{fig:passages}. The first scenario, Fig. \ref{fig:passages} (a), occurs in the case of a doorway or the openings and closings of a narrow passage. In such scenarios, the invisible humans exert equal forces from two different directions, which align the robot at the center of the passage, but it cannot pass through until the invisible constraint is switched off. However, the threat of invisible humans still exists, therefore, we make the robot act cautiously and move it with a lower velocity ($\leq$ \SI{0.3}{\meter\per\second}) until it passes through the passage. To detect this scenario, we use the positions of the invisible humans and the robot to check whether an \textit{isosceles triangle} is formed with the three vertices. The robot lies on the vertex, which connects the approximately equal sides, and humans are present at the base vertices. In order to limit false detections, we set some numerical limits on the lengths of the equal sides and the base. Assuming that a human has \SI{0.3}{\meter} radius and the robot has \SI{0.5}{\meter}, the length of the base should be $\geq$ \SI{1.6}{\meter}. When the clearance from obstacles or walls is taken into account it increases further. In this work, we set the limit on base length as \SI{3}{\meter}. Similarly, for the equal sides, there should be a minimum length of \SI{0.8}{\meter}, and we chose the limit to be \SI{2}{\meter}. These values are chosen empirically based on the tests in several situations. If the above conditions are satisfied, a passage is detected, and CoHAN switches to a new modality called \textit{Passing Through}, which sets the conditions mentioned above. 
The situation shown in Fig. \ref{fig:passages} (c) is almost the same as the doorway and we differentiate these scenarios by reading the center value of the laser scan data. If the value in the data is less than the length of the perpendicular bisector of the triangle's base, then it is identified as a pillar or large obstacle approach. CoHAN identifies this as a new modality, but for now, we handle it the same we handle the doorway.

The situation shown in Fig. \ref{fig:passages} (b) is different from the other two as the robot's passage is blocked by an invisible human on one side and obstacle clearance on the other side. As the robot, may or may not align in this case, it is handled differently. We check the angle of the laser scan corresponding to the detected corner and read the value of the data that is symmetrical to this angle along the direction of the robot. Finally, if the difference between the distance of this laser scan data and the invisible human from the robot's position is $<$ \SI{1}{\meter}, we identify it as a wall passage and set CoHAN to \textit{Pass Through} mode. The threshold of \SI{1}{\meter} is chosen empirically here.

\section{Experiments}
The proposed approach, after being completely integrated with CoHAN is tested in several settings. In this section, we show four interesting scenarios and present a detailed analysis. In all these experiments, we assume $h_{rad}=$\SI{0.3}{\meter} and set $k=10$ and $\alpha=0.2$. We use ROS-melodic with Ubuntu 18.04, and all the scenarios are simulated using MORSE \cite{echeverria2011modular} simulator. The simulated human agents used in the experiments are controlled using InHuS \cite{favier2021intelligent}, a human simulator developed in our lab.

\subsection{The effect of the Invisible Humans constraint}
\begin{figure}[h!]
    \centering
    \includegraphics[width=0.7\columnwidth]{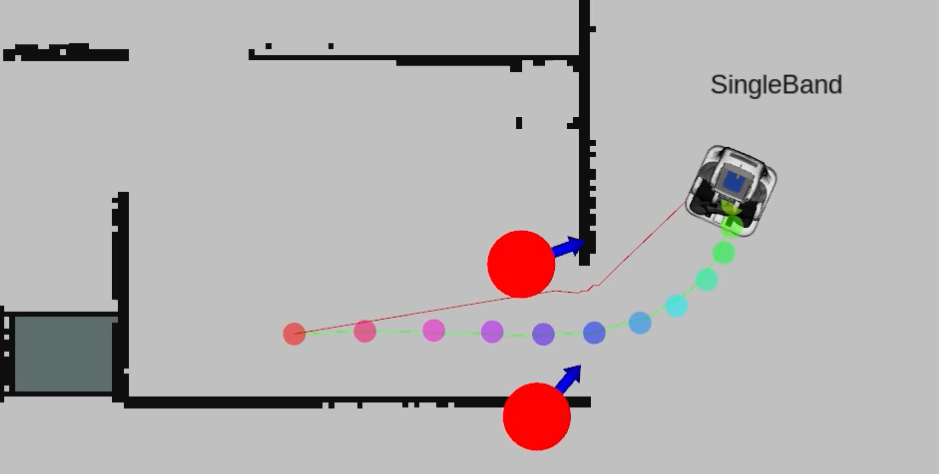}
    \caption{The robot passing through the door under the presence of invisible humans. The colored circles represent the poses of the robot and different color corresponds to different time instance.}
    \label{fig:door_scene}
\end{figure}

\begin{figure}[h]
% \begin{subfigure}{.5\columnwidth}
% %   \centering
%   % include first image
%   \includegraphics[width=0.9\textwidth]{figures/without_inv_door_path.png}
%   \caption{Path without constraint}
% \end{subfigure}
% \hspace{-0.17cm}
% \begin{subfigure}{.5\columnwidth}
% %   \centering
%   % include second image
%   \includegraphics[width=0.9\textwidth]{figures/inv_humans_door_path.png} 
%   \caption{Path with constraint}
% \end{subfigure}
% \begin{subfigure}{0.5\columnwidth}
% %   \centering
%   % include first image
%   \includegraphics[width=\textwidth]{figures/vel_without.png}
%   \caption{Speed without constraint}
% \end{subfigure}
% \hspace{-0.17cm}
% \begin{subfigure}{0.5\columnwidth}
% %   \centering
%   % include second image
%   \includegraphics[width=\textwidth]{figures/vel_with.png} 
%   \caption{Speed with constraint}
% \end{subfigure}
\centering
\includegraphics[width=0.85\columnwidth]{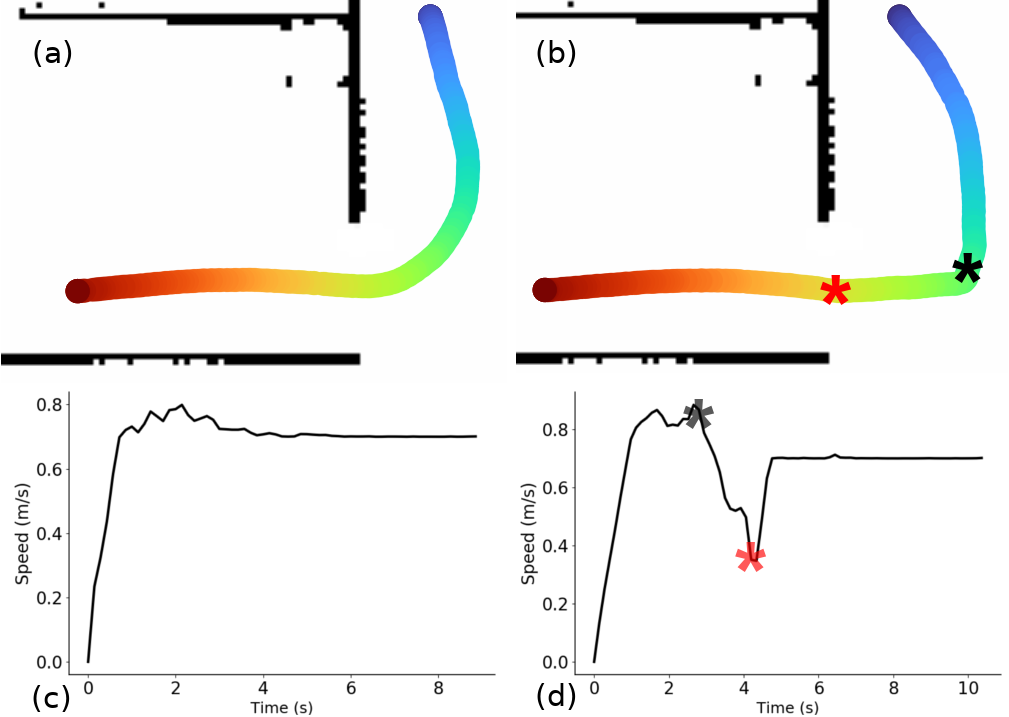}
\caption{Paths and speed profiles of the robot passing the door without (a, c) and with (b, d) the Invisible Humans constraint. The color of the paths indicates the time and progress of the robot, from blue to red (start to goal). In (a) the robot crosses the door “full speed”. In (b) it decelerates before entering the door (black star) and has the lowest speed at the entrance to the door (red star) around $4.2s$ corresponding to the shortest distance to invisible humans.}
\label{fig:door_constraint}
\end{figure}

To show the effect of introducing the invisible humans constraint into CoHAN, we present the robot with a door crossing scenario as shown in Fig. \ref{fig:door_scene}. We test scenarios without and with the invisible humans constraint and the corresponding paths of the robot are presented in Fig. \ref{fig:door_constraint} (a)  and Fig. \ref{fig:door_constraint} (b) respectively. The paths are colored, and the color moves from blue to red as the robot moves from start to goal. It can be clearly seen from these paths that the inclusion of the constraint made the robot more cautious as it takes a larger distance and aligns its path earlier to pass through the doorway. The corresponding speed plots are shown in Fig. \ref{fig:door_constraint} (c) and (d). Comparing the plot in Fig. \ref{fig:door_constraint} (d) with the profiles in Fig. \ref{fig:door_constraint} (c), it can be clearly seen that the robot slows down significantly before passing through the passage. The cautious behavior of the robot is again reflected in these speed profiles. 

% Useful for Subfigures
\subsection{Navigation in the presence of visible and invisible humans}
\begin{figure}[h!]
    \centering
%     \begin{subfigure}{0.9\columnwidth}
%     \centering
%     \includegraphics[width=0.9\columnwidth]{figures/corridor_scene_final.png}
%     % \caption{Corridor with many openings}
%     \end{subfigure}
%     \begin{subfigure}{0.9\columnwidth}
%   \centering
%   % include second image
%   \includegraphics[width=0.9\textwidth]{figures/pillar_scene_final.png} 
% %   \caption{Pillar Corridor}
% \end{subfigure}
\includegraphics[width=0.8\columnwidth]{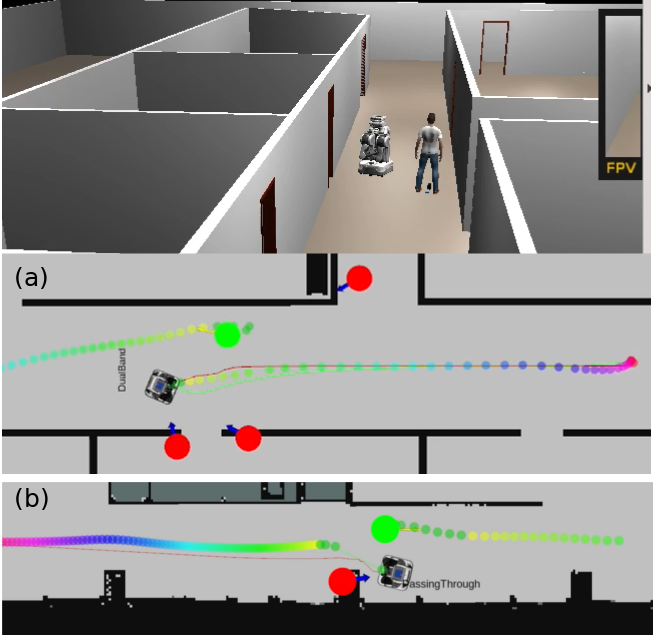}
    \caption{Corridor scenarios used for testing CoHAN. (a) Corridor with many openings where the robot continuously anticipates the emergence of humans. (b) Corridor with pillars between passage that creates complex navigation scenarios. In both these settings, the robot tries to find a balance between visible and invisible humans. The green circle is the visible human interacting with the robot, while the red circles are estimated invisible humans. The colored path with circles is the planned trajectory of the robot.}
        \label{fig:corridor_scene}
\end{figure}

The inclusion of the invisible humans into human-aware navigation planning should not cause discomfort to the visible humans that are moving around the robot. To show that CoHAN finds a fine balance between the invisible and visible humans, we present two corridor scenarios, one with many doors (or openings) as shown in Fig. \ref{fig:corridor_scene} (a) and the other with pillars as shown in Fig. \ref{fig:corridor_scene} (b). In both of these scenarios, the robot faces complex situations where it has to find a balance between the visible and the invisible humans.

% Useful for Subfigures
\begin{figure}[h!]
\begin{subfigure}{\columnwidth}
  \centering
  % include first image
  \includegraphics[width=0.8\textwidth]{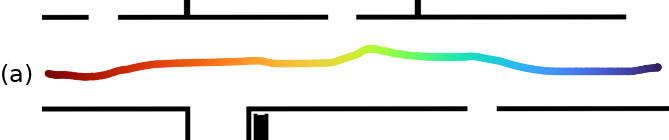}
%   \caption{Path: Many openings corridor}
\end{subfigure}
\hspace{-0.17cm}
\begin{subfigure}{\columnwidth}
  \centering
  % include second image
  \includegraphics[width=0.8\textwidth]{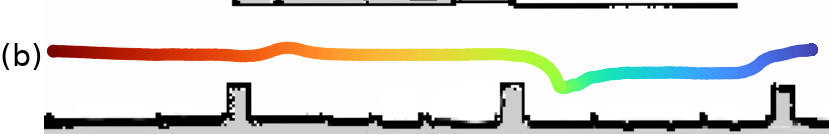} 
%   \caption{Path: Pillar Corridor}
\end{subfigure}
\begin{subfigure}{0.5\columnwidth}
  \centering
  % include first image
  \includegraphics[width=0.9\textwidth]{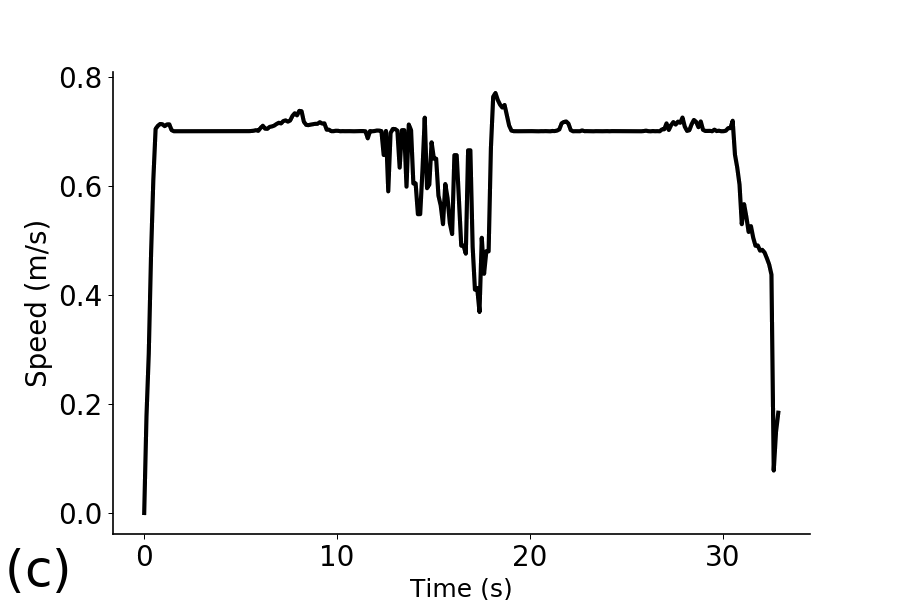}
%   \caption{Speed: Many openings corridor}
\end{subfigure}
\hspace{-0.17cm}
\begin{subfigure}{0.5\columnwidth}
  \centering
  % include second image
  \includegraphics[width=0.9\textwidth]{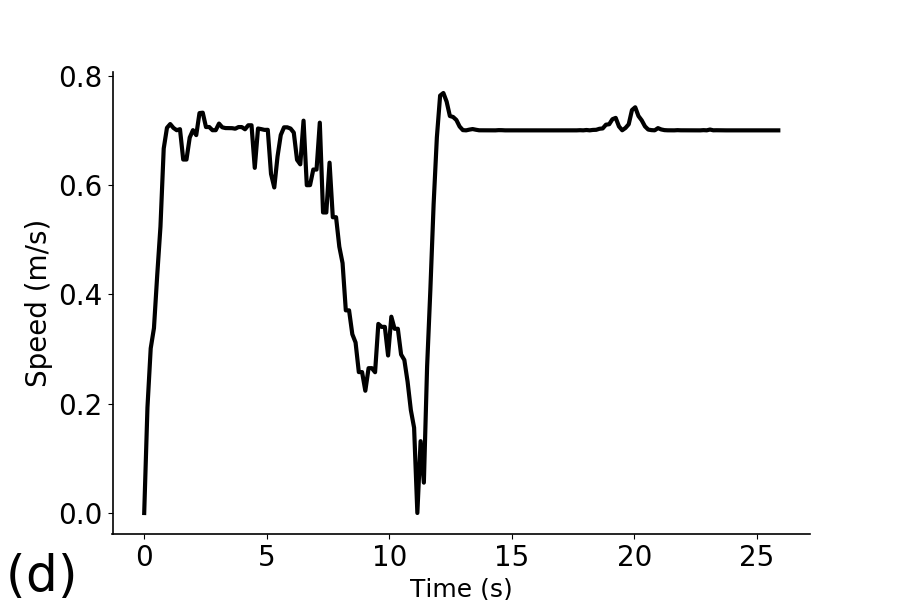} 
%   \caption{Speed: Pillar corridor}
\end{subfigure}
\caption{Paths and speeds profiles of the robot in corridor scenarios. (a), (c) correspond to the corridor with many openings. (b), (d) correspond to the pillar corridor. The color of the paths indicates the time and progress of the robot, from blue (start) to red (goal).}
\label{fig:corridors}
\end{figure} 
In the case of the corridor with many openings, the robot anticipates that a human might emerge anytime and tries to move away from the openings. However, when it sees a human passing through the corridor, it tries to provide more space to the human by moving to one side. At the same instance, it faces the forces from the invisible humans and tries to find a balance between these and the visible human. By observing the path and speed profile of this scenario from Fig. \ref{fig:corridors} (a) and (c), we can see that the robot moves away as well as reduces its speed rapidly to accommodate the visible human. Nonetheless, it does not move very close to the wall as it anticipates a human emergence. 
 
 In the corridor with pillars, the robot faces another complex situation where it has to pass through a very narrow opening and has to let the human pass through the same as shown in Fig. \ref{fig:corridor_scene} (b). From the path and speed profiles of this scenario from Fig. \ref{fig:corridors} (b) and (d), we can see that the robot slows down rapidly while moving to a side and momentarily stops before moving forward again. Here, the robot stops and lets the visible human pass before it can continue its navigation. Further, it detects the narrow passage scenario discussed in Section III and changes to \textit{Pass Through} mode. We can, therefore, infer that CoHAN always tries to find a fine balance between visible and invisible humans and can mitigate very complex situations.
 \subsection{Sudden emergence of a human}
 \begin{figure}[h]
    \centering
    \includegraphics[width=0.9 \columnwidth]{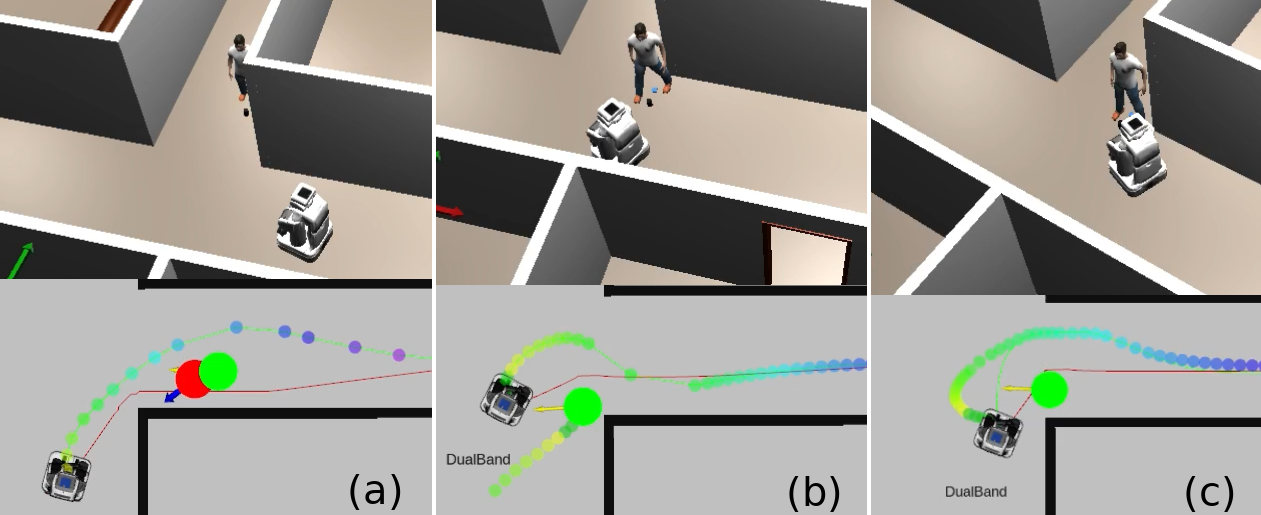}
    \caption{Sudden emergence scenario.  The colored paths with circles is the planned trajectory of the robot. (a) Shows the anticipated invisible human in red and the real human in green. The robot starts moving away from the corner. (b) The robot has seen the human and adjusted its trajectory to provide more space to the human. (c) The scenario without the invisible humans constraint. The robot moves very near to the wall blocks the human momentarily before adapting its path.}
    \label{fig:emergence}
\end{figure}
% The final scenario we discuss in this section shows a situation where a human emerges suddenly from an occluded region from where the robot is already anticipating an invisible human. The snapshots of this scenario before and after the emergence are shown in Fig. \ref{fig:emergence}. Before a human actually emerges from the corridor, a possible position of the invisible human is already estimated, and as shown Fig. \ref{fig:emergence} (a), it approximately overlaps with the real human. The robot starts moving away slowly because of this anticipation and suddenly a real human appears in front of it. The robot quickly adapts its trajectory and moves away from the human slowing down a little before continuing to its goal. From Fig. \ref{fig:appear_plots} (a), we can see the discontinuity in the path when the human emerges. However, the total change in path is not very drastic as the robot was already anticipating a human. The speed profile in Fig. \ref{fig:appear_plots} (b) an oscillation that occurred when the robot moved back suddenly and then slowed down until the human passes.
% From Fig. \ref{fig:appear_plots} (a), we can see the discontinuity in the path when the human emerges. However, the total change in path is not very drastic as the robot was already anticipating a human. The speed profile in Fig. \ref{fig:appear_plots} (b) an oscillation that occurred when the robot moved back suddenly and then slowed down until the human passes.
\begin{figure}[h!]
\centering
% \begin{subfigure}[t]{0.4\columnwidth}
% \hspace{-0.25cm}
%   \centering
%   % include first image
%   \includegraphics[width=0.7\textwidth]{figures/appear_new_traj_wo.png}
%   \caption{Path without constraint}
% \end{subfigure}
% \begin{subfigure}[t]{0.6\columnwidth}
% %   \centering
%   % include second image
%   \includegraphics[width=0.9\textwidth]{figures/appear_new_plot_wo.png} 
%   \caption{Speed without constraint}
% \end{subfigure}

% \begin{subfigure}[t]{0.4\columnwidth}
% \hspace{0.3cm}
% %   \centering
%   % include second image
%   \includegraphics[width=0.7\textwidth]{figures/appear_new_traj_with.png} 
%   \caption{Path with constraint}
% \end{subfigure}
% % \hspace{+0.17cm}
% \begin{subfigure}[t]{0.6\columnwidth}
% %   \centering
%   % include second image
%   \includegraphics[width=0.9\textwidth]{figures/appear_new_plot_with.png} 
%   \caption{Speed with constraint}
% \end{subfigure}
\includegraphics[width=0.8\columnwidth]{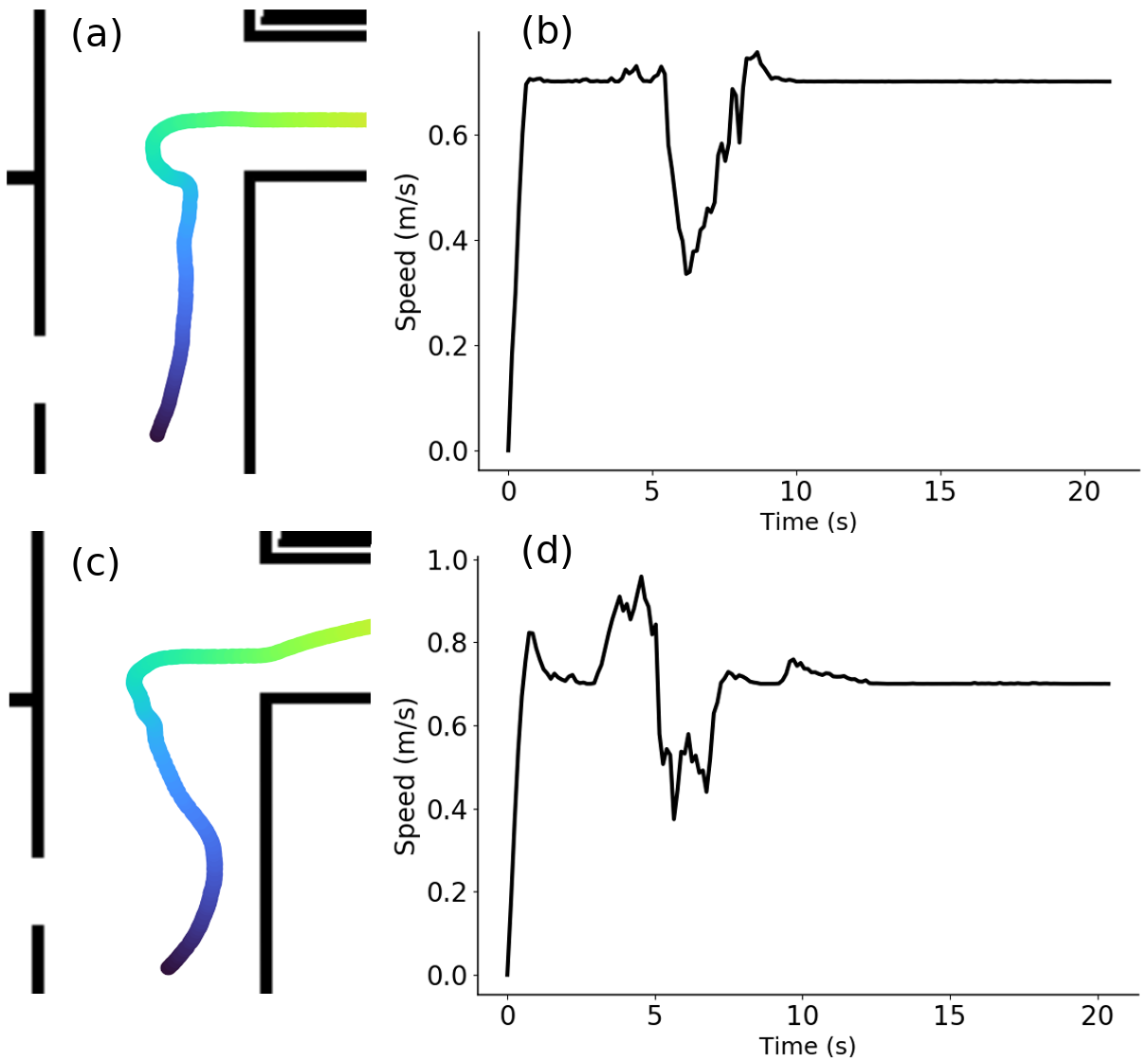}
\caption{Path and speed profiles of the sudden emergence scenario without (a, b) and with (c, d) the invisible humans constraint. The color of the path indicates the time and progress of the robot, from blue to green (start to goal).}
\label{fig:appear_plots}
\end{figure}
The final scenario we discuss in this section shows a situation where a human emerges suddenly from an occluded region. The snapshots of this scenario before and after the emergence are shown in Fig. \ref{fig:emergence}. The added invisible humans detection predicts a possible position of the human as shown in Fig. \ref{fig:emergence} (a), which approximately overlaps with the real human. The robot starts moving away from the wall slowly because of this anticipation, and suddenly a real human appears in front of it (Fig. \ref{fig:emergence} (b)). The robot quickly adapts its trajectory and moves away from the human, slowing down a little before continuing to its goal. However, without this detection, as shown in Fig. \ref{fig:emergence} (c), the robot moves close to the wall and blocks the human's way for a moment before adapting its path. The paths taken by the robot without and with the addition of invisible humans constraint to CoHAN are shown in Fig. \ref{fig:appear_plots} (a) and Fig. \ref{fig:appear_plots} (c) respectively. It is clear from these plots that the proposed constraint makes the robot move cautiously and lessens the surprise to humans. Further, the path of the robot is smoother in Fig. \ref{fig:appear_plots} (c) when compared to the path in Fig. \ref{fig:appear_plots} (a) as there are no sudden path changes. 

The speed profile in Fig. \ref{fig:appear_plots} (b) shows a sudden drop in the velocity of the robot. This occurs because CoHAN adapts its speed to prevent a possible collision and shock to the human. Then, the robot slowly moves away and plans a new path to the goal. The speed profile in Fig. \ref{fig:appear_plots} (d) is completely different compared to the last one. The robot starts drifting away from the wall (both x and y velocities) before seeing the human and this shows the increase in the velocity. When the human appears, it slows down and then changes its direction before continuing to the goal with almost a constant speed. This is the sharp change (slowdown) we see in Fig. \ref{fig:appear_plots} (d). 
\subsection{Quantitative Analysis}
\begin{figure}[h!]
\begin{subfigure}[t]{0.5\columnwidth}
  \centering
  % include first image
  \includegraphics[width=0.8\textwidth]{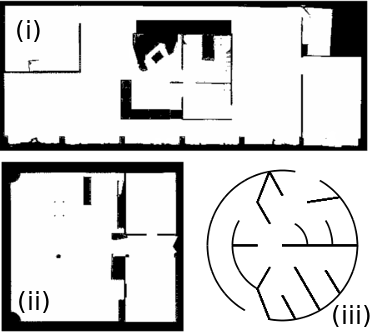}
  \caption{Maps used for testing (i) LAAS (ii) Bremen (iii) Random Maze}
\end{subfigure}
\hspace{-0.17cm}
\begin{subfigure}[t]{0.5\columnwidth}
%   \centering
  % include second image
  \includegraphics[width=0.9\textwidth]{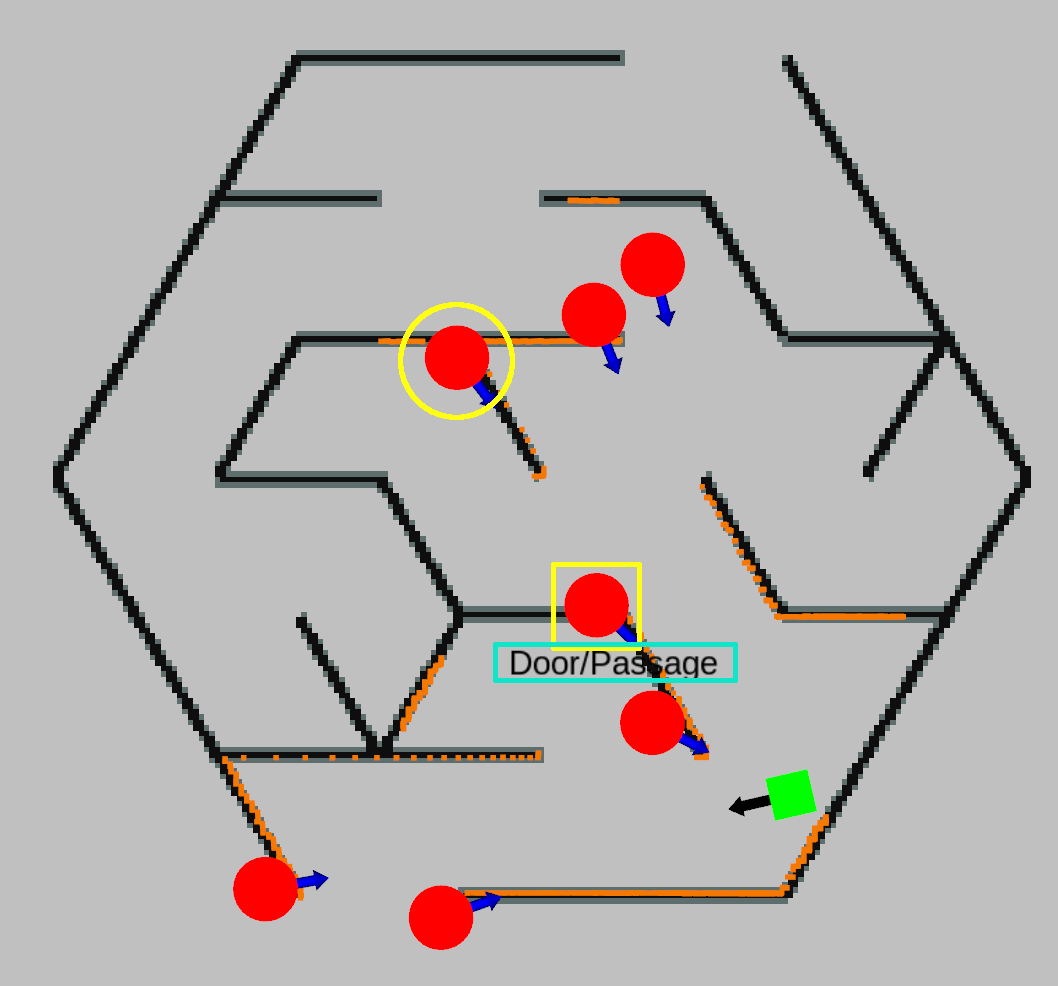} 
  \caption{Failed detections}
\end{subfigure}
\caption{(a) Different maps used for testing the robustness and accuracy of the approach. The map in (iii) shows an example of randomly generated maze. The other maps are standard ones. (b) Different types of failed detections. The detection in yellow circle is a false positive whereas that in yellow square is an overlap (possibly true). The passage detection is also wrong as the wall is detected as passage. The green cube with black arrow shows the robot and its direction. The red cylinders are the invisible humans.}
\label{fig:random_exps}
\end{figure}
For testing the proposed algorithm, we have designed randomized experiments. We either generate a random map using the Maze Generator\footnote{\url{https://github.com/razimantv/mazegenerator}} or randomize the position of the robot in the known map. The maps used for these experiments are shown in Fig. \ref{fig:random_exps} (a). The LAAS and Bremen maps are collected from the models of the real spaces. Fig. \ref{fig:random_exps} (a) (iii) shows a random map generated using the Maze Generator. Fig. \ref{fig:random_exps} (b) shows some failed detections using the proposed algorithm that are taken into account while calculating the accuracy. The red circles with blue arrows are the predicted invisible humans. The invisible human in the yellow circle is classified as a false positive as no human could be located inside the wall. The detection in the yellow square is similar, but it is not completely inside the wall. We call this case an `overlap' and classify this also as a false positive. The door/passage detected (cyan rectangle) in this picture is wrong, and we classify this as a false positive while calculating accuracy for passage detection. The green square with black arrow is the robot in the figure.
\subsubsection{Robustness and accuracy }
To test the robustness of the proposed algorithm, we perform several randomized experiments in different settings. In the LAAS and Bremen maps, we randomize the position of the robot and evaluate the detections manually. We did 50 such evaluations for each of the above two maps. In the next set of experiments, we generate a random map and place the robot in a random pose and then evaluate the detections. In this case, we have done 100 evaluations using 100 randomly generated maps. The calculated accuracy of our invisible humans detection algorithm based on these 200 experiments is $76.85\%$. However, if we include the overlaps are true detections, it increases by over $12\%$ to $89.16\%$. These overlaps could be reduced by improving the filtering. Table \ref{acc} shows the list of experiments and the accuracy in each case.
\begin{table}[h!]
    \centering
    \begin{tabular}{|c|c|c|}
    \hline
    & \multicolumn{2}{c|}{Invisible Humans Detection}\\
    \cline{2-3}
    Experiment & \textit{Accuracy ($\%$)}& \textit{Accuracy with overlap ($\%$)} \\
    \hline
    \textit{LAAS} & 91.82 & 94.55\\
    \hline
    \textit{Bremen} & 65.28 & 81.94 \\
    \hline
    \textit{Random} & 76.90 & 90.42 \\
    \hline
     \textbf{Overall} & \textbf{76.85} & \textbf{89.16} \\
    \hline
    \end{tabular}
    \caption{Accuracy calculated based on 200 experiments in 3 different environments. By correcting the overlapping detections, the accuracy could be increased by over $12\%$.}
    \label{acc}
\end{table}
\vspace{-0.5cm}
\begin{table}[h!]
    \centering
    \begin{tabular}{|c|c|c|}
    \hline
    & \multicolumn{2}{c|}{Passage Detection}\\
    \cline{2-3}
    Experiment & \textit{Accuracy ($\%$)}& \textit{Misses due to limits ($\%$)} \\
    \hline
    \textit{LAAS} & 66.00 & 30.00 \\
    \hline
    \textit{Bremen} & 54.00 & 36.00 \\
    \hline
    \textit{Random} & 65.00 & 14.0 \\
    \hline
     \textbf{Overall} & \textbf{62.50} & \textbf{23.50} \\
    \hline
    \end{tabular}
    \caption{Accuracy calculated based on 200 experiments in 3 different environments. Increasing the limits of detection could increase the accuracy by over $23\%$. These limits could be decided based on the requirement.}
    \label{acc_pass}
\end{table}

For calculating the accuracy of passage detection, we have performed similar experiments as above and evaluated the detections in 200 experiments. Here, we classify the detection simply either as true or false. There are also cases where there are no detections. In such cases, no detection within limits is classified as false, and out of the limits is classified as a miss. Table \ref{acc_pass} shows the accuracy of passage detection in different settings. The overall accuracy based on these experiments is around $62.50\%$. Note that the percentage of misses is around $23.50\%$. This can be improved by having higher or adaptive detection limits. They have to be set based on the requirement.

\subsubsection{Comparison with other planners}
To test the advantage of the proposed approach, we compare the sudden emergence scenario using three different planners. The first one is Simple Move Base (SMB), where humans are added using the laser scan. Then we use CoHAN with and without the proposed constraint as the other two planners.
\begin{table}[ht!]
    \centering
    \begin{tabular}{|l|c|}
    \hline
    \textbf{Planners} & \textbf{Min. Human-Robot Distance ($m$)}\\
    \hline
      \textit{SMB} & 0.584\\
    \hline
     \textit{CoHAN} & 0.922\\
    \hline
     \textit{CoHAN with constraint} & 1.247\\
    \hline
    \end{tabular}
    \caption{Minimum human-robot distance in the sudden emergence scenario using 3 different planners. The results are the average over 5 runs.}
    \label{compare_table}
\end{table}
As the proposed work makes the robot cautious and tries to reduce the surprise to humans, we check the minimum human-robot distance while navigating using these planners. We have performed 5 runs of the same scenario with each planner, and the results are presented in Table \ref{compare_table}. From Table \ref{compare_table}, we can see that adding the invisible humans constraint to CoHAN makes the robot maintain a larger distance from the human around a corner. Keeping distance from humans avoids surprise to humans and provides time for the robot planner to adapt slowly.

\section{Real World Results}
\begin{figure}[h!]
    \centering
    \includegraphics[width=0.8\columnwidth]{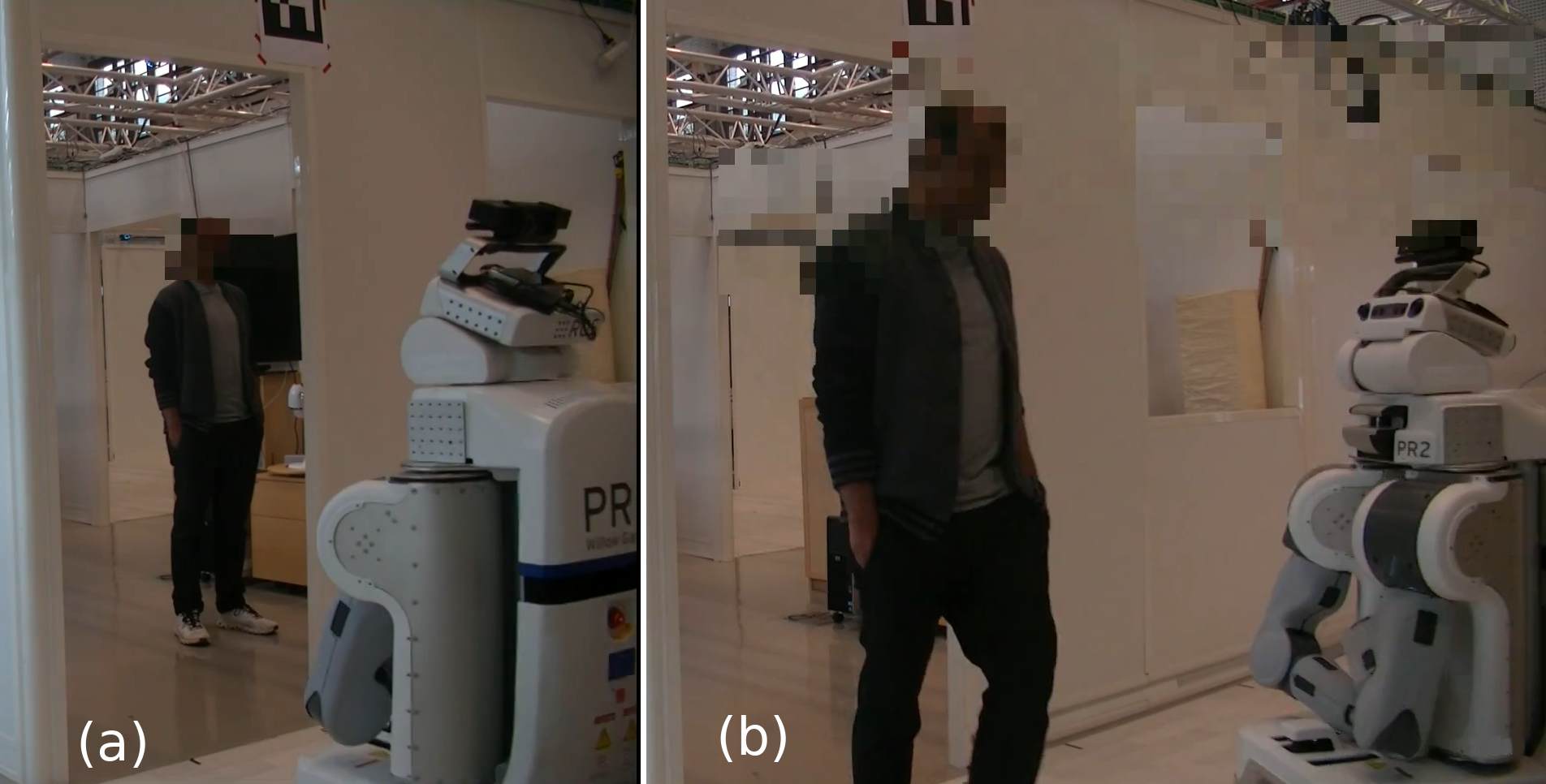}
    \caption{Real world experiment setting (a) Human is inside the room and does not move. (b) The human starts coming out of the door as the robot approaches the door. In both scenarios, the robot tries to pass through the door.}
    \label{fig:real_photo}
\end{figure}
\begin{figure}[h!]
    \centering
    \includegraphics[width=0.8\columnwidth]{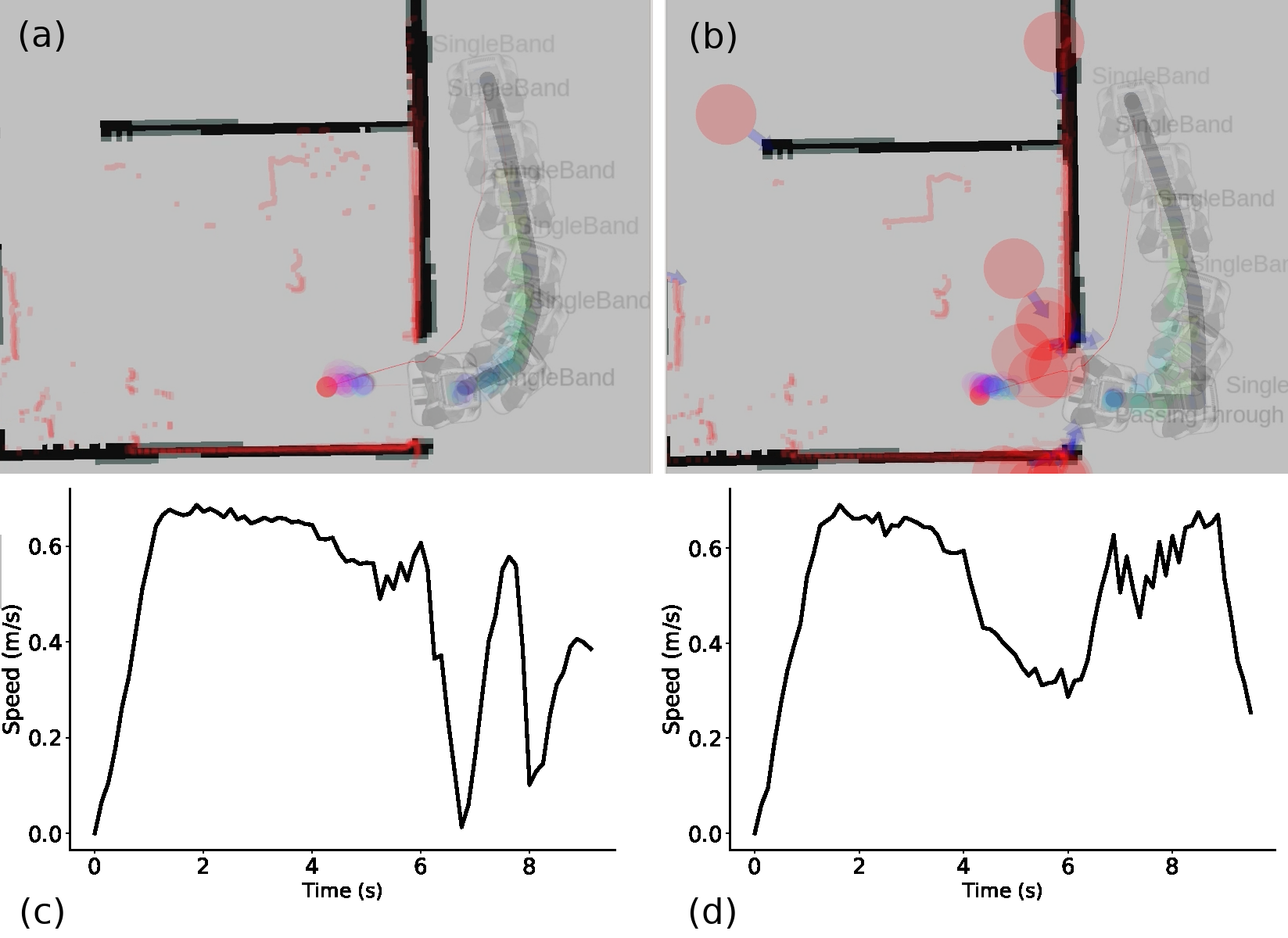}
    \caption{Paths and speed profiles without (a, c) and with (b, d) invisible humans constraint. The addition of the constraints makes robot takes larger turn (b).}
    \label{fig:real_plots}
\end{figure}
\begin{figure}[h!]
\centering
% \begin{subfigure}[t]{0.5\columnwidth}
%   \centering
%   \includegraphics[width=0.8\textwidth]{figures/move_final_new.png}
%   \caption{Path in sudden appear}
% \end{subfigure}
% \hspace{-0.17cm}
% \begin{subfigure}[t]{0.5\columnwidth}
%   \includegraphics[width=0.8\textwidth]{figures/move_final_new_plot.png} 
%   \caption{Speed profile in sudden appear}
% \end{subfigure}
\includegraphics[width=0.8\columnwidth]{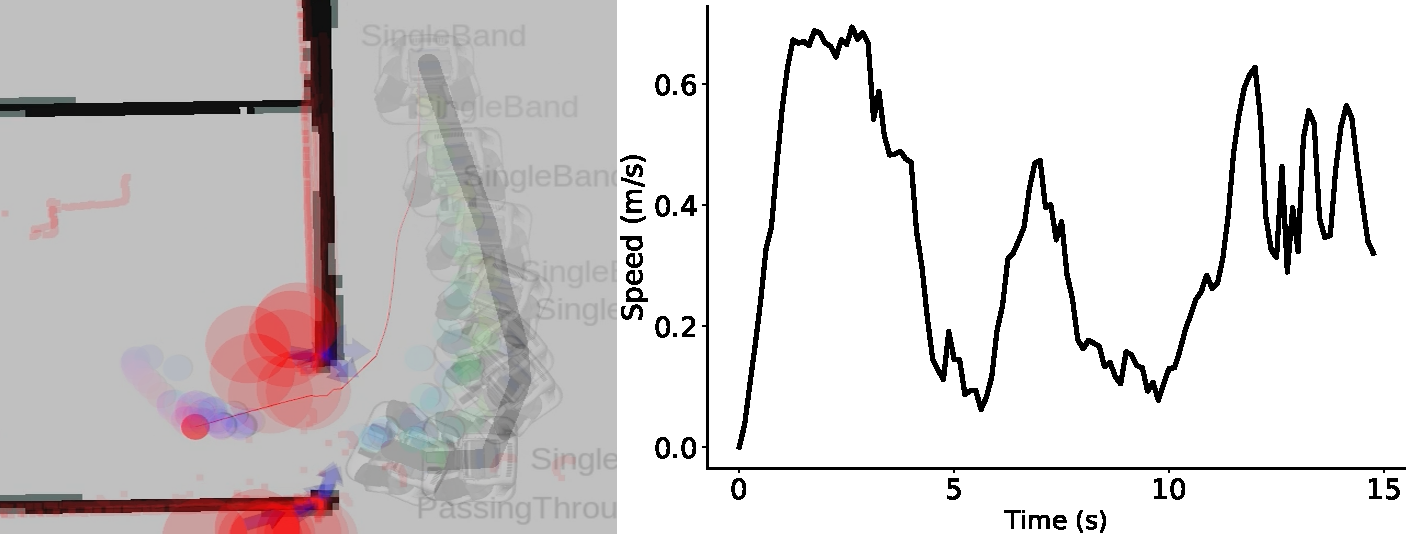}
\caption{Path and speed profile in the sudden emergence scenario.}
\label{fig:real_move_plots}
\end{figure}

The CoHAN system is installed on the PR2 robot in our lab and then tested in the doorway scenario discussed above. We performed two kinds of experiments, as shown in Fig. \ref{fig:real_photo}. In the first situation, shown in Fig. \ref{fig:real_photo} (a), the human remains stationary, whereas in the second situation, shown in Fig. \ref{fig:real_photo} (b), he comes out of the door as the robot approaches the door. The first situation is tested, with and without the invisible humans constraint, and the results are shown in Fig. \ref{fig:real_plots}. We can see from the figure that the real-world results match the results of the simulation approximately both in the paths and the speed profiles. Note that in Fig. \ref{fig:real_plots} (c), the robot halts momentarily. This may be because of a sudden human appearance or moving very close to the wall.
The second situation is similar to the sudden emergence scenario, and the results of this run are shown in Fig. \ref{fig:real_move_plots}. The robot takes a larger turn and slows down twice, once to align itself for the door and then when the human emerges. The video showing some more experiments and the tested scenarios can be found at {\small{\url{https://youtu.be/cbeFRkEdGgA}}}.

\section{Discussion and Limitations}
The introduction of invisible humans into human-aware navigation planning is relatively new and requires further research. We present one possible approach to address this issue. What is particularly interesting here is that our approach is modeled as a situation assessment and prediction ability to integrate into a mobile robot human-aware navigation. Having this, we will have a robot that can interact, using several modalities, with humans present in its field of view while making provisions to adapt to humans that are not yet seen. One difficulty we faced is the integration all these features without being ``too conservative" and avoiding another case of the ``freezing" robot. However, there are still some limitations to this approach. Since the approach is based on a 2D map, we can have false detections in the regions visible through the head of the robot but not through the base. There may also be some false detections in complex maps. Both these can be mitigated by augmenting the current approach with new sensor data and filtering further. 

% The second limitation is the imperfect detection of the invisible humans, where sometimes, the detected human overlaps with an obstacle. 
% The presented algorithm needs to be refined further and the current work is a preliminary step.

\section{Conclusion}
We have proposed an approach to estimate the locations of invisible humans on a 2D map. These invisible humans are then integrated into our human-aware navigation planner via a new constraint. We also show how narrow passages can be identified by exploiting the detected invisible humans. We have presented the qualitative analysis of several simulated scenarios, followed by the results of accuracy and comparisons with some planners. Finally, we have shown the real-world experiments and presented some discussion. In the future, we plan to refine this approach further and address the different modalities identified in a better manner. We also aim to build a complete human-aware navigation system that can address very intricate human-robot interactions.

\bibliographystyle{ieeetr}
\bibliography{references.bib}

\end{document}